%% file: main.tex
\documentclass[10pt,twocolumn,letterpaper]{article}

\usepackage{cvpr}
\usepackage{times}
\usepackage{epsfig}
\usepackage{graphicx}
\usepackage{amsmath}
\usepackage{amssymb}
\usepackage{svg}
\usepackage{booktabs}
\usepackage{multirow}
\usepackage{caption} 
\usepackage{enumitem}
\usepackage[pagebackref=true,breaklinks=true,colorlinks,bookmarks=false]{hyperref}

\definecolor{mypink}{rgb}{0.858, 0.188, 0.478}

\makeatletter
\def\blfootnote{\xdef\@thefnmark{}\@footnotetext}
\makeatother

\newcommand{\mat}[1]{\mathbf{#1}}

\cvprfinalcopy 

\def\cvprPaperID{689} 

\ifcvprfinal\fi
\begin{document}

\title{Correspondence Networks with Adaptive Neighbourhood Consensus}

\author{Shuda Li$^1$\thanks{indicates equal contribution}\hspace{2em} Kai Han$^2$\footnotemark[1]\hspace{2em} Theo W. Costain$^1$\hspace{2em} Henry Howard-Jenkins$^1$\hspace{2em} Victor Prisacariu$^1$\vspace{1em}\\
$^1$Active Vision Lab \& $^2$Visual Geometry Group\\
Department of Engineering Science, University of Oxford\\
{\tt\small \{shuda, khan, costain, henryhj, victor\}@robots.ox.ac.uk}
}

\maketitle
\blfootnote{\textcopyright\ 2020 IEEE. Personal use of this material is permitted. Permission from IEEE must be obtained for all other uses, in any current or future media, including reprinting/republishing this material for advertising or promotional purposes, creating new collective works, for resale or redistribution to servers or lists, or reuse of any copyrighted component of this work in other works.}

\begin{abstract}
In this paper, we tackle the task of establishing dense visual correspondences between images containing objects of the same category.
This is a challenging task due to large intra-class variations and a lack of dense pixel level annotations.
We propose a convolutional neural network architecture, called adaptive neighbourhood consensus network (ANC-Net), that can be trained end-to-end with sparse key-point annotations, to handle this challenge.
At the core of ANC-Net is our proposed non-isotropic 4D convolution kernel, which forms the building block for the adaptive neighbourhood consensus module for robust matching.
We also introduce a simple and efficient multi-scale self-similarity module in ANC-Net to make the learned feature robust to intra-class variations.
Furthermore, we propose a novel orthogonal loss that can enforce the one-to-one matching constraint.
We  thoroughly evaluate the effectiveness of our method on various benchmarks, where it substantially outperforms state-of-the-art methods. 



\end{abstract}


\begin{figure*}[!th]
\centering
  \includegraphics[width=0.89\linewidth]{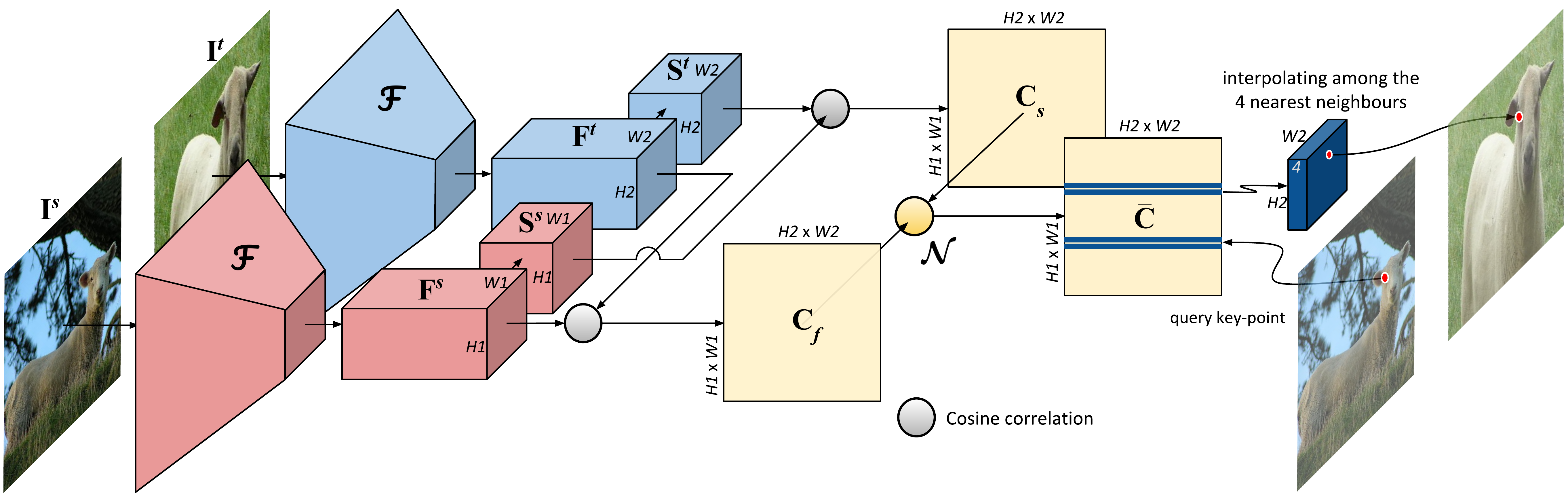}
  \caption{\textbf{An overview of ANC-Net}. Given a pair of images ($\mathbf{I}^s$, $\mathbf{I}^t$), ANC-Net can predict their pixel-wise semantic correspondences. A CNN backbone $\mathcal{F}$ first extracts features $\mathbf{F}^s$ and $\mathbf{F}^t$. Our multi-scale self-similarity module then captures the self-similarity features $\mathbf{S}^s$ and $\mathbf{S}^t$ based on $\mathbf{F}^s$ and $\mathbf{F}^t$. We can then obtain $\mathbf{C}_s$ from $\mathbf{S}^s$ and $\mathbf{S}^t$, and $\mathbf{C}_f$ from $\mathbf{F}^s$ and $\mathbf{F}^t$. Taking $\mathbf{C}_f$ and $\mathbf{C}_s$ as input, our ANC module $\mathcal{N}$ will predict a refined $\bar{\mathbf{C}}$, from which the pixel-wise correspondences can be retrieved with interpolation.
  }
\label{fig:overview}
\end{figure*}
\input{introduction}
\input{related_work}

\input{method}
\input{experiment}

\input{conclusion}

\noindent\textbf{Acknowledgements.} We gratefully acknowledge the support of the European Commission Project Multiple-actOrs Virtual EmpathicCARegiver for the Elder (MoveCare) and the EPSRC Programme Grant Seebibyte  EP/M013774/1.

{\small
\bibliographystyle{ieee_fullname}
\bibliography{reference}
}

\end{document}

%% file: introduction.tex
\section{Introduction}
Establishing visual correspondences has long been a fundamental problem in computer vision. 
It has seen variety of applications in areas such as 3D reconstruction~\cite{Agarwal_ICCV09_BuildRome,Sattler_CVPR17_Benchmark_6DOF}, 
image editing~\cite{HaCohen11Non}, scene understanding~\cite{Liu_PAMI11_SIFT}, and object detection~\cite{Duchenne_ICCV11}.

Earlier works mainly focused on estimating correspondences for images of the same scene or object (\ie instance-level correspondences) using hand-crafted features such as SIFT~\cite{Lowe2004SIFT} or HOG~\cite{Dalal05HOG}. 
Recently, finding correspondences for different instances from the same category (\ie semantic correspondences) has attracted more and more attention\cite{Choy_NIPS16_UCN,han2017scnet,Rocco_NIPS18_NCNet,Huang_ICCV19_DCCNet,min2019hyperpixel}. 
In this paper, we focus on the problem of establishing dense correspondences for a pair of images depicting different instances from the same category.
This task is extremely challenging due to large intra-class variation in properties such as colour, scale, pose, and illumination.
Further, it is unreasonably expensive, if not impossible, to provide dense annotations for such image pairs. 

To deal with the challenges mentioned above, we introduce a convolutional neural network (CNN), called Adaptive Neighbourhood Consensus Network (ANC-Net), which can produce reliable semantic correspondences without requiring dense human annotations.
ANC-Net takes a pair of images as input and predicts a 4D correlation map, containing the matching scores for all possible matches between the two images.
The most likely matches can then be retrieved by finding the matches giving the maximum matching scores. 

ANC-Net consists of a CNN feature extractor, a multi-scale self-similarity module, and an adaptive neighbourhood consensus module.
At the core of ANC-Net is our proposed non-isotropic 4D convolution, which incorporates an adaptive neighbourhood consensus constraint for robust matching, and our proposed multi-scale self-similarity module, which aggregates multiple self-similarity features, which are insensitive to intra-class appearance variation\cite{Kim_CVPR17_FCSS}. 

CNN features have been very popular for the task of correspondence estimation due to their promising performance, and most state-of-the-art methods are based on CNN features ~\cite{Rocco_NIPS18_NCNet,min2019hyperpixel,Huang_ICCV19_DCCNet,Kim_CVPR17_FCSS,Choy_NIPS16_UCN}.
Like other methods, ANC-Net also extracts features with a pre-trained CNN. 
However, instead of directly using the CNN features to calculate matching scores, we introduce the multi-scale self-similarity.
Self-similarity has been introduced in existing methods~\cite{Huang_ICCV19_DCCNet,Kim_CVPR17_FCSS}.
Unlike other methods that either use self-similarity as an extra feature alongside raw CNN features~\cite{Huang_ICCV19_DCCNet}, or use computationally expensive irregular self-similarity patterns~\cite{Kim_CVPR17_FCSS}, our self-similarity features are both computationally cheap to obtain, and do not need combining with raw CNN features, whist still capturing the complex self-similarity patterns.

With reliable feature representation, the matching scores for each individual feature pair can then be calculated.
However, as the individual feature pairs do not contain any matching validity information, matching by direct feature comparison can be rather noisy.
To mitigate this, correspondence validity constraints should be applied to obtain reliable matching scores.
Neighbourhood consensus, which measures how many pairs are matched in the neighbourhoods of the two points under consideration, turns to be one of the most effective correspondence validity constraints, and has been successfully introduced in recent work~\cite{Rocco_NIPS18_NCNet,Huang_ICCV19_DCCNet}.
However, \cite{Rocco_NIPS18_NCNet} and \cite{Huang_ICCV19_DCCNet} assume neighbourhoods of the same size for the two points in consideration.
Unfortunately, this assumption does not hold in practice, as objects in real images typically have different scales and shapes.
Therefore, adopting neighbourhoods of the same size is very likely to be affected by unrelated neighbours (\eg background parts).
To address this issue, we propose an adaptive neighbourhood consensus module, which can select the correct neighbourhoods. 

As mentioned earlier, the cost of labelling ground truth means fully supervised learning with dense annotations is not feasible.
Instead, our model can effectively make use of sparse key-point annotations. To enforce the one-to-one mapping constraint, which is crucial for plausible correspondences, we further propose a novel one-to-one mapping loss, called orthogonal loss, to regularise the training.

To summarise, our contributions are four fold:
\setlist{nolistsep}
\begin{itemize}[noitemsep]
    \item We introduce ANC-Net for the task of dense semantic correspondence estimation, which can be trained with sparse key-point annotations.
    \item We propose a non-isotropic 4D convolutional kernel, which forms the building block for the adaptive neighbourhood consensus module for robust matching. 
    \item We propose a simple and efficient multi-scale self-similarity to make the feature matching robust to intra-class variation. 
    \item We propose a novel orthogonal loss that can enforce the one-to-one matching constraint, encouraging plausible matching results.
\end{itemize}
We thoroughly evaluate the effectiveness of our method on various benchmarks, where it substantially outperforms state-of-the-art methods. 
Our code can be found at~\url{https://ancnet.avlcode.org/}. 

%% file: related_work.tex
\section{Related work}

The semantic correspondence estimation problem is often considered as either a pixel-wise matching problem, an image alignment problem, or a flow estimation problem.
Earlier works used hand-crafted features, such as SIFT~\cite{Lowe2004SIFT} or HOG~\cite{Dalal05HOG}, to establish semantic correspondences~\cite{Liu_PAMI11_SIFT,Kim2013dsp,hur2015generalized,ham2018proposal_flow,Ham16PF,taniai2016joint}.
Here, we briefly review recent CNN based methods. 

\noindent
\textbf{Pixel-wise matching.}\enspace
Long et al.~\cite{Long2014doconv} transferred the features pre-trained on an image classification task to pixel-wise correspondence estimation.
Choy et al.~\cite{Choy_NIPS16_UCN} introduced a method to learn a feature embedding for the correspondence problem, by pulling positive features pairs close and pushing negative feature pairs away.
Han et al.~\cite{han2017scnet} proposed a CNN model that tries to match image patches considering both appearance and geometry information, and obtains the pixel-wise correspondences by interpolation.
Novotny et al.~\cite{Novotn2018selfsuper} introduced a method to learn geometrically stable features with self-supervised learning by applying a synthetic warp to the images.
More recently, Rocco et al.~\cite{Rocco_NIPS18_NCNet} proposed to construct a CNN model that incorporates neighbourhood consensus information to refine the 4D tensor storing all the matching scores, which are obtained from pre-trained CNN features.
Huang et al.~\cite{Huang_ICCV19_DCCNet} introduced a method to incorporate self-similarity based on~\cite{Rocco_NIPS18_NCNet} and fuse different features with an attention mechanism.
Min et al.~\cite{min2019hyperpixel} showed that effectively combining features extracted from different layers can provide significant benefits for the dense semantic correspondence estimation task.

\noindent
\textbf{Image alignment.}
Rocco et al.~\cite{Rocco17} developed a CNN architecture that can predict the global geometric transformation between two images by training on synthetically warped data.
Seo et al.~\cite{paul2018attentive} improved~\cite{Rocco17} by introducing attention based offset-aware correlation kernels.
Rocco et al.~\cite{Rocco_CVPR18_WeakAlign} presented an end-to-end trainable CNN architecture that uses weak image-level supervision, which is trained by a soft inlier counting loss in a similar spirit to RANSAC.
Jeon et al.~\cite{Jeon2018parn} introduced a hierarchical learning procedure to progressively learn affine transformations to align the images in a chaos-to-fine manner.
Kim et al.~\cite{Kim_NIPS18_RTNet} introduced to a recurrent transformer network, which is trained with an iterative process and can predict the transformations between a pair of images. 

\noindent
\textbf{Flow estimation.}
Fischer et al.~\cite{fischer2015flownet} introduced an end-to-end trainable model called FlowNet, which is trained on synthetic data to predict optical flow.
FlowNet is further improved by Ilg et al.~\cite{IMSKDB17} in several aspects. 
Kim et al.~\cite{Kim_CVPR17_FCSS} proposed a learnable self-similarity feature, which is then used to estimate an dense affine transformation flow for each feature location.
The semantic correspondences can then be obtained by applying such transformations.
Lee et al.~\cite{Lee19} introduced a method to use images annotated with binary foreground masks, and subjected to synthetic geometric deformations, to train a CNN model with a mask consistency loss and a flow consistency loss.
Besides these, there are also some methods that learn the flow using videos~\cite{Wang_cvpr19_learning,Lai_BMVC19} by considering temporal consistency.





%% file: method.tex
\section{Method}
Given a pair of images ($\mathbf{I}^s$, $\mathbf{I}^t$), our objective is to find pixel-wise correspondences between the two images.
We propose a CNN, ANC-Net, which takes ($\mathbf{I}^s$, $\mathbf{I}^t$) as input and produces a 4D correlation map containing the matching scores for all possible pairs in the feature space of the two images.
Pixel-wise correspondence then can be extracted by interpolation among the most likely matches in the feature space.
The model can be trained with a supervised loss on sparse key-point annotations in an end-to-end manner.
To encourage one-to-one matching, we propose using a novel loss, called the orthogonal loss, together with the supervised loss on sparse key-point annotations, for training our model. 

Figure~\ref{fig:overview} illustrates the main architecture of our network.
It consists of a feature extractor $\mathcal{F}$, a multi-scale self-similarity module,  and an adaptive neighbourhood consensus (ANC) module $\mathcal{N}$. 
The feature extractor $\mathcal{F}$ is composed of a sequence of standard convolutional layers.
We first feed the two images into $\mathcal{F}$, and get a pair of feature maps $\mathbf{F}^s$ and $\mathbf{F}^t$. 
The multi-scale self-similarity module $\mathcal{S}$ consists of two convolutional layers followed by a concatenation operation to fuse them into the multi-scale features. 
With $\mathbf{F}^s$ and $\mathbf{F}^t$, $\mathcal{S}$ will produce the multi-scale self-similarity feature maps $\mathbf{S}^s$ and $\mathbf{S}^t$ which capture the complex self-similarity patterns. 
We can then obtain the 4D correlation map $\mathbf{C}_s$ from $\mathbf{S}^s$ and $\mathbf{S}^t$, and the 4D correlation map $\mathbf{C}_f$ from $\mathbf{F}^s$ and $\mathbf{F}^t$.
However, $\mathbf{C}_s$ and $\mathbf{C}_f$ are often noisy as they lack the constraints to enforce the correspondence validity, and thus are unreliable for directly extracting correspondences. 
Our proposed ANC module $\mathcal{N}$, which is realised with a stack of non-isotropic 4D convolutions, takes $\mathbf{C}_s$ and $\mathbf{C}_f$ as input, refining them by considering neighbourhoods with varying sizes.
Finally, the ANC module combines the refined correlation maps by simply summing up the two, producing a single 4D correlation map $\bar{\mathbf{C}}$, from which reliable correspondences can be retrieved.
$\mathbf{C}_s$ is introduced to capture the second order (and higher) cues derived from the raw features. $\mathbf{C}_s$ shares a similar structure to $\mathbf C_f$, allowing both to be refined using a neighbourhood consensus module without introducing extra learnable parameters.
Experiments show that the proposed self-similarity module outperforms similar methods~\cite{Kim_CVPR17_FCSS, Huang_ICCV19_DCCNet}.

In this section, we will first introduce the multi-scale self-similarity module in Section~\ref{sec:mss}. We then, in Section~\ref{sec:anc}, describe the adaptive neighbourhood consensus matching validity module. Section~\ref{sec:map} will discuss the approach to enforcing global constraints over the output of the neighbourhood consensus by maximising an a posteriori estimation. Finally, we describe the learning objectives for training our network in Section~\ref{sec:loss}.


\subsection{Multi-scale self-similarity}
\label{sec:mss}

\begin{figure}[t]
\centering
  \includegraphics[width=0.9\linewidth]{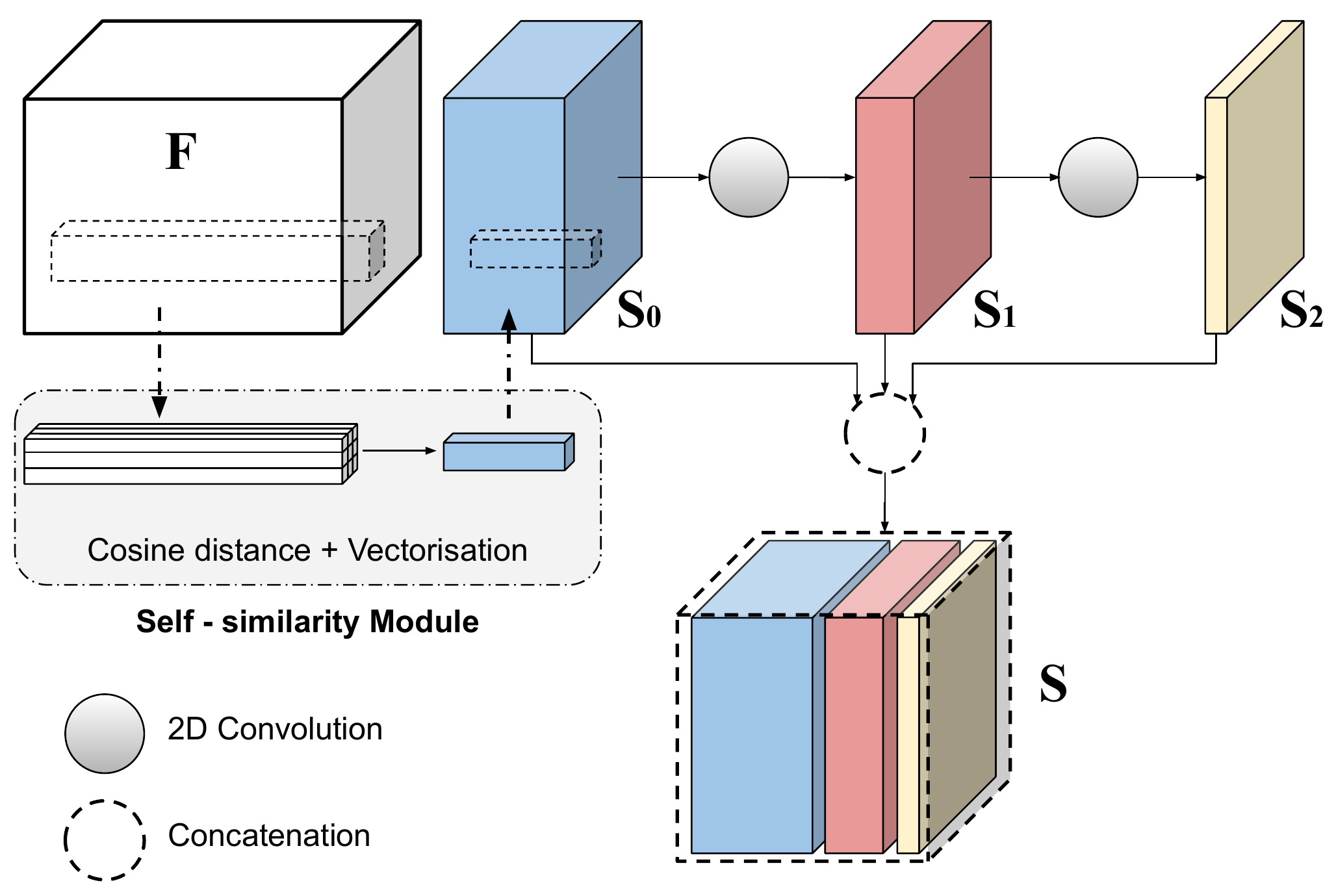}
\caption{\textbf{Self-similarity module}. The top left figure illustrates the calculation of a self-similarity score over the 3 $\times$ 3 window. Specifically, the cosine distances between each of the 9 features and the centre feature are calculated and then vectorised into the $\mathbf{S}_0$. In the bottom, we first calculate the $\mathbf{S}_0$ from the feature map $\mathbf F$, and then perform two levels of 2D convolutions, each followed by an activation function (ReLU) to produce $\mathbf{S}_1$ and $\mathbf{S}_2$. Finally, the initial similarity score $\mathbf{S}_0$, its first scale filtered features  $\mathbf{S}_1$, and second filtered features  $\mathbf{S}_2$ are concatenated together to form final feature map $\mathbf S$.}
\label{fig:pss}
\end{figure}

Self-similarity has been shown to be effective for the task of semantic correspondence estimation~\cite{Kim_CVPR17_FCSS, Huang_ICCV19_DCCNet}. 
Given a feature map $\mathbf{F} \in \mathbb{R}^{h_f \times w_f \times d}$ established by the backbone feature extractor, a self-similarity map measures the local similarity pattern at each feature location.
One way to extract the self-similarity feature for the feature vector $\mathbf f_{ij}$ at $(i, j)$ in $\mathbf{F}$ is to calculate the cosine distance between itself and its neighbours.
Figure~\ref{fig:pss} illustrates the self-similarity module when considering the $3\times 3$ neighbours of a given feature vector.
This approach results in 9 self-similarity scores for each $\mathbf f_{ij}$.
We further vectorise each of the $3\times 3$ self-similarity scores into a 9-vector, which make up the self-similarity feature map $\mathbf{S}_0 \in \mathbb{R}^{h_f \times w_f \times 9}$.

To further capture the correlations among different self-similarity features,
we apply two 2D convolutional layers with zero padding on $\mathbf S_0$.
Given the output feature maps for the two layers are $\mathbf{S}_1$ and $\mathbf{S}_2$, we then concatenate the 3-scales $\mathbf{S}_0$, $\mathbf{S}_1$, and $\mathbf{S}_2$ together to form an enhanced feature map $\mathbf{S}$, which will serve as the input to the later layers.
With the feature maps $\mathbf{S}^s$ and $\mathbf{S}^t$ of source and target images respectively, we can obtain the 4D correlation map $\mathbf{C}_s$. 


Unlike DCCNet~\cite{Huang_ICCV19_DCCNet}, where the self-similarity of a single scale is considered, and the self-similarity scores are then concatenated with $\mathbf F$ and convolved using a point-wise convolution which is intended to use the self-similarity to re-weight the raw features, our method avoids fusing with $\mathbf F$ to reduce redundancy, as the features in $\mathbf F$ have already been implicitly included in $\mathbf{S}_0$.
Further, we extract more complex self-similarities than DCCNet and make use of multi-scale self-similarities to bootstrap the features.
Thus, we capture more complex features from a much larger local window as well as second order (and higher) information.

As will be shown in the experiments, our multi-scale self-similarity module performs better than that of DCCNet.
It is also worth noting that FCSS~\cite{Kim_CVPR17_FCSS} proposes a similar design, however their self-similarity score is defined using a set of irregular point pairs within the local window which is more complex to implement.
In contrast, we adopt the design of correlating the centre feature with neighbours for simplicity and computation efficiency, and as a result, our simplified self-similarity module outperforms FCSS in all benchmarks.

Both $\mathbf C_f$ and $\mathbf C_s$ are complementary to each other as we hypothesise they are dominated by first order and higher order cues respectively.
They will be refined by the following ANC module independently and then combined.

\subsection{Adaptive neighbourhood consensus}
\label{sec:anc}
\begin{figure}[t]
\begin{center}
  \includegraphics[width=0.48\textwidth]{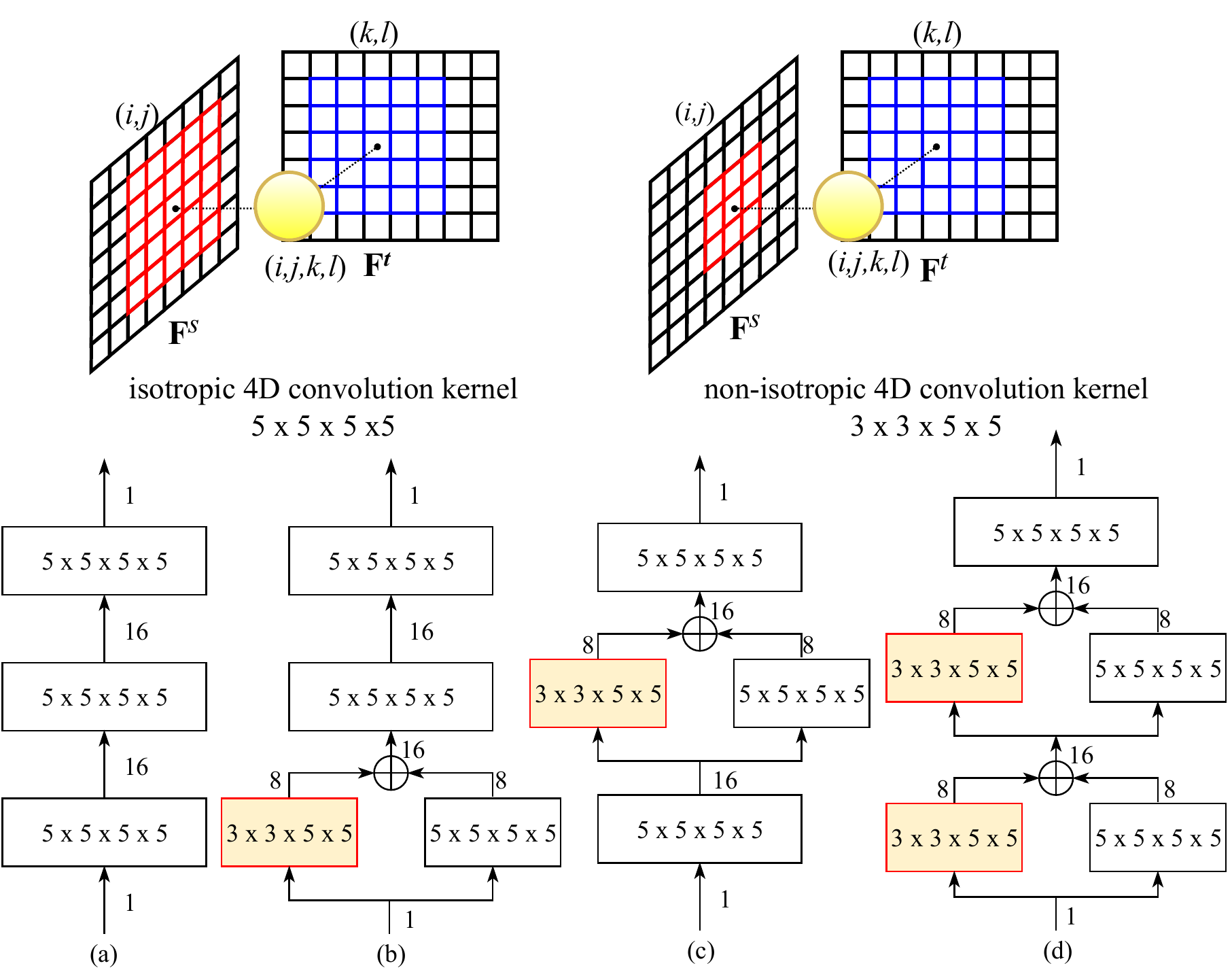}
\end{center}
  \caption{\textbf{Adaptive neighbourhood consensus}. The top row illustrates an isotropic and a non-isotropic 4D convolutional kernel. The bottom row illustrates the architecture of (a) the non-isotropic in NC-Net~\cite{Rocco_NIPS18_NCNet}  and (b-d) three ANC candidates. $\oplus$ denotes concatenation of feature maps. The numbers $\{1, 16, 16, 1\}$ denote the input and output channels for the 4D kernels. 
  The non-isotropic 4D convolutions are always zero padded so that the size of the 4D correlation remains the same size after each convolution. 
  } 
\label{fig:anc}
\end{figure}

Neighbourhood consensus has been shown to be effective for filtering the noisy 4D correlation map~\cite{Rocco_NIPS18_NCNet,Huang_ICCV19_DCCNet}.
Multiple layers of the \emph{isotropic} 4D convolutional kernels, \ie kernels with identical size in each dimension, are applied on the 4D correlation map to refine it. The isotropic 4D convolution with size $5\times 5 \times 5 \times 5$ is illustrated in top left of Figure~\ref{fig:anc}. It can be seen that the kernel establishes two neighbourhoods with the same size for both images. However, objects in real images often have varying scales and shapes, therefore, two neighbourhoods depicting the same semantic meaning are very likely to have different sizes. Thus, using neighbourhoods of the same size for both images may introduce noise (\eg unrelated background) when determining a match.

To deal with the problem, we introduce the adaptive neighbourhood consensus (ANC) module which contains a set of \emph{non-isotropic} 4D convolutional layers. As illustrated in the top right of Figure~\ref{fig:anc}, the non-isotropic 4D convolution has dimensions of $3\times 3 \times 5\times 5$, defining the neighbourhood of $3\times 3$ and $5\times 5$. 

To handle objects in real images with varying scales and shapes, we can combine our \emph{non-isotropic} 4D kernels with \emph{isotropic} 4D kernels so that the model can dynamically determine which set of convolutions should be activated to handle objects of various sizes. 
We consider 3 candidate architectures (shown in Figure~\ref{fig:anc}) in our experiments with each non-isotropic 4D convolution using zero padding. 
Unless stated otherwise, we use (d) in our experiments, as it gives the best performance in our evaluation.
This is possibly because (d) allows for more scale variation than the others.
This choice might ignore better designs than (d), but the main point in this work is to demonstrate the effectiveness of the ANC module.

It is also worth noting that it is unnecessary to have both $p \times p \times q \times q$ and $q \times q \times p \times p$ kernels in the model where $p$ and $q$ are the sizes of some kernel dimensions, as the bidirectional neighbourhood consensus filter in Eq.~\ref{eq:anc} (which will be explained next) effectively tries both the configurations of small vs large neighbourhood and large vs small by reversing the matching direction, and the effect of both filters are equivalent due to the bidirectional matching. 


Let $\mathcal{N}$ be the module of our adaptive neighbourhood consensus. It takes a 4D correlation map $\mathbf{C}_s$ or $\mathbf{C}_f$ as input and refining them.
Their refined counterparts can then be combined to form $\bar{\mathbf C}$.
We apply $\mathcal{N}$ to both matching directions (\ie matching $\mathbf{I}^s$ to $\mathbf{I}^t$ and matching $\mathbf{I}^t$ to $\mathbf{I}^s$), so that our model is invariant to the order of the images.
More importantly, this allows $\mathcal{N}$ to only include one $p\times p\times q\times q$ non-isotropic kernel to handle the small to large as well as the large to small neighbourhood.
In particular, the refined 4D correlation map can be obtained by 

\begin{equation}
\bar{\mathbf{C}}=\mathcal{N}(\mathbf{C}_s)+\left(\mathcal{N}\left(\mathbf{C}_s^\top\right)\right)^\top + \mathcal{N}(\mathbf{C}_f)+\left(\mathcal{N}\left(\mathbf{C}_f^\top\right)\right)^\top,
\label{eq:anc}
\end{equation}
where $^\top$ denotes the swapping of the matching direction given an image pair, \ie, $(C^\top)_{ijkl} = C_{klij}$.

\subsection{Most likely matches}
\label{sec:map}
After obtaining the refined 4D correlation map $\bar{\mathbf{C}}$, we follow~\cite{Rocco_NIPS18_NCNet} to apply soft mutual nearest neighbour filtering, \ie, for each $\bar{c}_{i j c d}$ in $\bar{\mathbf{C}}$, we replace it by $\hat{c}_{i j c d} = r_{i j k l}^{s} r_{i j k l}^{t} \bar{c}_{i j k l}$ where $r_{i j k l}^{s}=\frac{\bar{c}_{i j k l}}{\max _{a b} \bar{c}_{a b k l}}$ and $r_{i j k l}^{t}=\frac{\bar{c}_{i j k l}}{\max _{c d} \bar{c}_{i j c d}}$, which downweights the scores of matches that are not mutual nearest neighbours. Next, we perform soft-max normalisation to the scores $\hat{c}_{i j k l}$.
The normalised scores can be interpreted as the matching probabilities.
In particular, the probability of a given point at $(i, j)$ in $\mathbf{I}^s$ being matched with an arbitrary point  $(k, l)$ in $\mathbf{I}^t$ is
\begin{equation}
v_{i j k l}^{t}=\frac{\exp \left(\hat{c}_{i j k l}\right)}{\sum_{c d} \exp \left(\hat{c}_{i j c d}\right)} .
\end{equation}
Similarly, the probability of a given point at $(k, l)$ in $\mathbf{I}^t$ being matched with an arbitrary point  $(i, j)$ in $\mathbf{I}^s$ is
\begin{equation}
v_{i j k l}^{s}=\frac{\exp \left(\hat{c}_{i j k l}\right)}{\sum_{a b} \exp \left(\hat{c}_{a b k l}\right)} .
\end{equation}
For a given position $(i, j)$ in $\mathbf{I}^s$,  the most likely match $(k, l)$ in $\mathbf{I}^t$ can be found by 
\begin{equation}
(k, l) = \underset{c d}{\arg \max }\ v_{i j c d}^{t} .
\label{eq:kl_v}
\end{equation}
Similarly, for a given position $(k, l)$ in $\mathbf{I}^t$, the most likely match $(i, j)$ in $\mathbf{I}^s$ can be found by 
\begin{equation}
(i, j) = \underset{a b}{\arg \max } v_{a b k l}^{s} .
\label{eq:ij_v}
\end{equation}
After retrieving the correspondences in the feature space with Eq.~\ref{eq:kl_v} and Eq.~\ref{eq:ij_v}, the pixel-wise correspondences can be obtained by interpolation. 

\subsection{Learning objective}
\label{sec:loss}
For the tasks of establishing dense semantic correspondences, it is impossible to obtain dense ground-truth labelling for all training image pairs due to the huge amount of human labour required. 
In practice, one can easily label only a few key-points of the objects in an image.
These key-points often indicate the objects parts with concrete semantic meaning (\eg eyes, mouths, body joints, etc.).
Sparse key-point annotations are included in many existing datasets including PF-PASCAL~\cite{ham2018proposal_flow}, Spair-71k~\cite{min2019hyperpixel}, CUB~\cite{Welinder_2010_CUB} and others.
There are also other forms of alternative annotations, such as image level pairwise annotations~\cite{Rocco_NIPS18_NCNet,Huang_ICCV19_DCCNet}, or object masks~\cite{Lee19}.
In this paper, we are interested in the sparse key-point annotations, as they are more directly linked to our objective to learn semantic correspondences.



The sparse key-point annotations provide a straightforward way to train a CNN model for semantic matching, in which we minimise the distances between features of matched key-points (\eg \cite{Choy_NIPS16_UCN}).
However, this is not applicable to ANC, because the feature space ANC operates is a 4D correlation map, rather than a 3D feature map consisting of per pixel feature vectors. Therefore, we introduce a simple but effective supervised loss on 4D correlation maps to train our model. 

\begin{figure}[th]
\centering
  \includegraphics[width=0.95\linewidth]{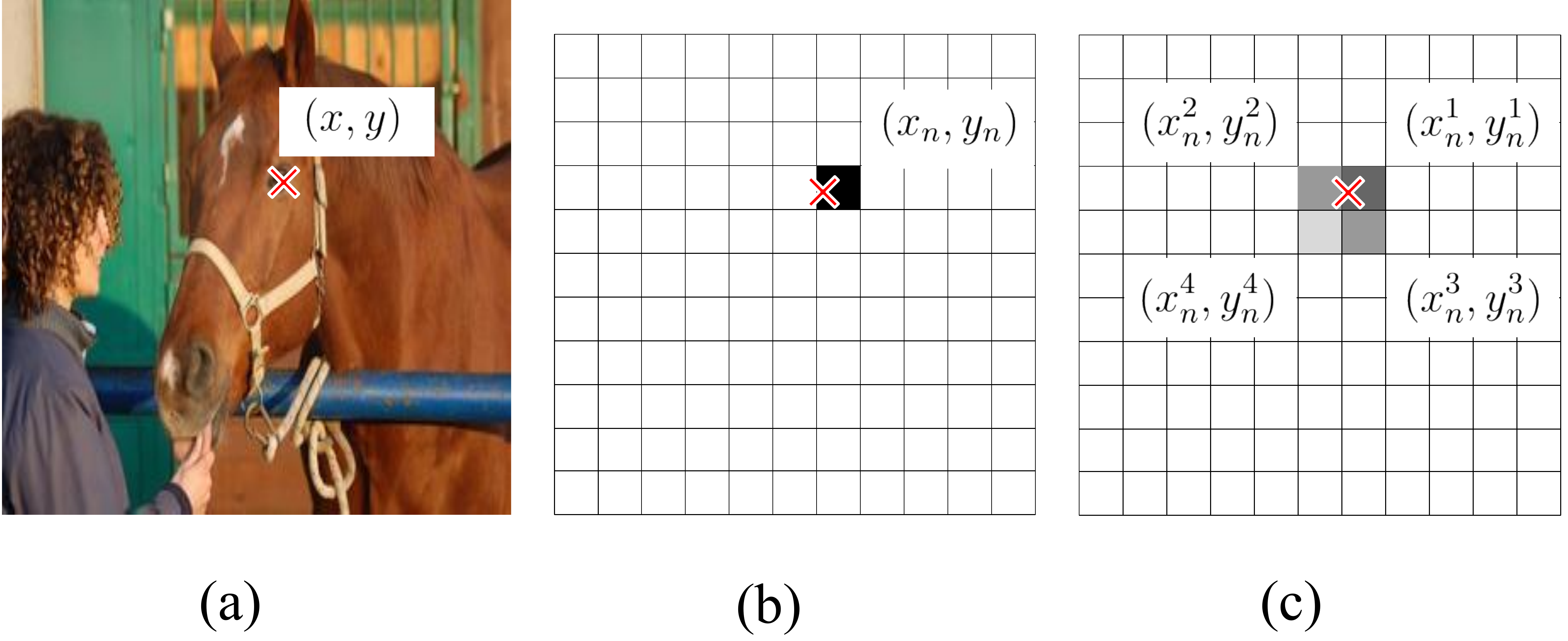}
   \caption{\textbf{Generating the ground-truth probability map for each key point}. (a) The key point $(x, y)$ is a key point in the image coordinates. (b) $(x_n, y_n)$ is the nearest neighbour of $(x_c, y_c)$ which is re-scaled coordinate $(x,y)$ to the feature map resolution. (c) $(x_n^1, y_n^1)$,  $(x_n^2, y_n^2)$,  $(x_n^3, y_n^3)$, and  $(x_n^4, y_n^4)$ are the four nearest neighbours to $(x_c, y_c)$. }
\label{fig:1nn}
\end{figure}

For each key-point $(x, y)$ in the image (\eg~Figure~\ref{fig:1nn}(a)), 
we first re-scale $(x, y)$ to the same resolution as the feature map, giving the re-scaled coordinates $(x_c, y_c)$.
Since $(x_c, y_c)$ is a sub-pixel coordinate, it can not be used as the target in the feature map directly.
Instead, we can simply pick the nearest neighbour $(x_n, y_n)$ of $(x_c, y_c)$ in the feature map to be the target (see Figure~\ref{fig:1nn} (b)).
However, this will introduce errors caused by ignoring the offset between the $(x_n, y_n)$ and $(x_c, y_c)$.
As the resolution of the feature map is much smaller than that of the image, small offsets in the feature map will cause large errors in the image.
To compensate for the offset, we take the four nearest neighbours into consideration (see Figure~\ref{fig:1nn} (c)), rather than the single nearest neighbour.
In particular, we pick the four nearest neighbours $(x_n^1, y_n^1)$,  $(x_n^2, y_n^2)$,  $(x_n^3, y_n^3)$, and  $(x_n^4, y_n^4)$, and set scalar values $t_1$, $t_2$, $t_3$, and $t_4$ to them representing the probability of being the considered as target.
$t_1$, $t_2$, $t_3$, and $t_4$ are proportional to their distances to $(x_c, y_c)$, and $\sum_{j=1}^{4} t_j = 1$. We then apply 2D Gaussian smoothing on the four nearest neighbour probability map obtained above. We found that such smoothing can effectively enhance the performance.
In this way, each key-point location annotation is converted into a 2D probability map. Next, we reshape the smoothed 2D probability map into a $(h_c \times w_c)$-vector for the key-point $(x, y)$, followed by $L_2$ normalisation.
For the source image $\mathbf{I}^s$ containing $n$ key-points, we can therefore construct its target as a matrix $\mathbf{M}_{gt} \in \mathbb{R}^{ n\times (h_c \times w_c)}$ with each row being a probability vector of a ground truth matching key-point in the target image $\mathbf{I}^t$. Let $\mathbf{M}_{gt}$ and $\mathbf{M}$ be the ground truth and prediction. Note that $\mathbf{M}$ can be obtained by flattening the first two and last two dimentions of $\bar{\mathbf{C}}$ (after mutual nearest neighbour filtering), and taking the same $n$ rows corresponding to $\mathbf{M}_{gt}$. The loss function is then the Frobenius Norm between them for both matching directions:

\begin{equation}
\mathcal{L}_k = \|\mathbf{M}^s - \mathbf{M}^s_{gt}\|_F + \|\mathbf{M}^t - \mathbf{M}^t_{gt}\|_F,
\label{eq:L_k}
\end{equation}
where $\mathbf{M}^s$ denotes target probability map from $\mathbf{I}^s$ to $\mathbf{I}^t$ and $\mathbf M^t$ denotes inverse direction. 



\subsection{Enforcing one-to-one matching}
The one-to-one mapping (\ie one point can be only matched to one other point) turns out to be a useful clue for improve the matching accuracy in classic graph matching (GM)\cite{Zanfir_CVPR18_DeepGM,Jiang_AAAI17_NOGM}, which aims to match two given point sets (graphs) in two images.
Ideally, for our semantic correspondence estimation task, the result should also agree with the one-to-one mapping constraint.
This is especially helpful when there exist some repetitive patterns in the image (\eg a building with multiple identical windows).
GM methods always assume that the number of  key-points in two images are exactly the same.
However, this is often not the case in real applications.
For example, due to pose variation, some key-points may be visible in one image, but not in the other.
In this case, there exist one-to-none mappings in both images.
A plausible one-to-one matching constraint should be able to ignore the one-to-none matches in the data automatically.
To handle this problem, we introduce a novel loss, named the orthogonal loss, as it was inspired by the non-negative orthogonal GM algorithm~\cite{Jiang_AAAI17_NOGM}.
The idea is that when $\mathbf M\mathbf M^\top$ is an identity matrix $\mathbf I$, each row of $\mathbf M$ contains only one element, and the rest are zero, so we include a difference between $\mathbf M\mathbf M^\top$ and $\mathbf I$ in the loss. However, in reality, $\mathbf M$ may contain zero rows for one-to-none case. Therefore, our orthogonal loss term can be defined as 
\begin{equation}
    \mathcal{L}_o =  \|\mat{M}\mat{M}^\top - \mat{M}_{gt}\mat{M}_{gt}^\top\|_F ,
\end{equation}
where $\|.\|_F$ is a the Frobenius norm.  It is worth noting that $\mat{M}_{gt}\mat{M}_{gt}^\top$ has zeros on its diagonal that allows both one-to-one and one-to-none matches to be accurately penalised.
The orthogonal loss has to be combined with Eq.~\ref{eq:L_k} as it has no impact over the prediction accuracy.
It simply regularises the model by encouraging one-to-one predictions.
The overall loss of our model can be written as 
\begin{equation}
    \mathcal{L} =  \mathcal{L}_k +\alpha \mathcal{L}^m_o, 
\end{equation}
where $\alpha$ is a weight balancing term, which is set to $0.001$ in all our experiments, and $\mathcal{L}^m_o = \|\mat{M}^s{\mat{M}^s}^\top - \mat{M}^s_{gt}{\mat{M}^s_{gt}}^\top\|_F + \|\mat{M}^t{\mat{M}^t}^\top - \mat{M}^t_{gt}{\mat{M}^t_{gt}}^\top\|_F$ by considering both matching directions.


%% file: experiment.tex
\section{Experimental results}

\subsection{Datasets and implementation details}
\noindent
\textbf{Datasets.} We evaluate our method on four public datasets, namely, PF-PASCAL~\cite{ham2018proposal_flow},  Spair-71k~\cite{min2019hyperpixel}, and CUB~\cite{Welinder_2010_CUB}. 
PF-PASCAL contains 1351 image pairs, which is approximately divided into 700 pairs for training 300 pairs for validation and 300 pairs for testing~\cite{han2017scnet,Rocco_NIPS18_NCNet}. 
Spair-71k dataset is much more challenging than the others as it contains both large view point differences and scale differences. We use the 12,234 pairs of testing pairs. Spair-71k is only used to evaluate the transferrability of the models trained on the PF-PASCAL training split. 
The CUB dataset contains 11,788 images of various species of birds with large variation of appearance, shape and pose. We randomly sample about 10,000 pairs from the CUB training data and test using the 5,000 pairs selected by~\cite{Krause_CVPR15_FineGrained}.

\noindent
\textbf{Implementation details.}
Our ANC-Net is implemented in the PyTorch~\cite{paszke2017automatic} framework.
We experiment with three convolutional networks as feature backbones, namely, ResNet-50, ResNet-101 and ResNeXt-101.
All of them are pre-trained on ImageNet~\cite{Li_CVPR09_ImageNet}, and the parameters are fixed during the training of our ANC-Net. 
The size of the self-similarity window is set to $5\times5$,
and channels of ANC module are set to $\{1, 16, 16, 1\}$.
The model is initially trained for 10 epochs using an Adam optimiser~\cite{Kingma_ICLR15_Adam} with a learning rate of $0.001$ and applying  Gaussian smoothing with a kernel size of 5 for ground truth probability map generation. 
The model is then fine-tuned for 5 epochs applying Gaussian smoothing with a kernel size of 3 followed by another 5 epochs with a kernel size of 0. 
To compare with DCCNet~\cite{Huang_ICCV19_DCCNet}, we implemented it based on the publicly available official implementation of NC-Net~\cite{Rocco_NIPS18_NCNet}.
Our implementation slightly surpassed the reported accuracy in~\cite{Huang_ICCV19_DCCNet}.
We also implemented UCN$_\text{ResNet-101}$ based on the publicly available official code~\cite{Choy_NIPS16_UCN}. 

\noindent
\textbf{Evaluation metric.}
Following common practice, we use the percentage of correct key-points (PCK@$\alpha$) as our evaluation metric. We report the results under PCK threshold $\alpha=0.1$.
$\alpha$ is set w.r.t. max($w_r, h_r$) where $w_r$ and $h_r$ are the width and height of either the image or the object bounding box. 
Following existing works~\cite{han2017scnet,Rocco_NIPS18_NCNet, Lee2019SFNet,min2019hyperpixel}, we use $\alpha$ w.r.t. the image size on PF-PASCAL, and w.r.t. the object bounding box on CUB and Spair-71k.

\begin{table}[htb]
\centering
\caption{\textbf{Comparison with state-of-the-art methods.}}
\label{tab:main}
\resizebox{0.9\linewidth}{!}{
\begin{tabular}{@{}lrrrr@{}}
	\toprule
	Methods                                                 & \multicolumn{1}{c}{PF-PASCAL}        & \multicolumn{1}{c}{CUB}  & \multicolumn{1}{c}{Spair-71k} \\\midrule\addlinespace[0.5em] 
	Identity mapping                                        & 37.0                         & 14.6             & 3.7             \\ \midrule                                         
	UCN$_\text{GoogLeNet}$~\cite{Choy_NIPS16_UCN}              & 55.6                         & 48.3             & 15.1             \\ 
	UCN$_\text{ResNet-101}$~\cite{Choy_NIPS16_UCN}              & 75.1                          & 52.1             & 17.7               \\ 
	SCNet$_\text{VGG-16}$~\cite{han2017scnet}                & 72.2                          & -                & -                \\

	Weakalign$_\text{ResNet-101}$~\cite{Rocco_CVPR18_WeakAlign} & 74.8                         & -                & 21.1             \\ 
	RTNet$_\text{ResNet-101}$~\cite{Kim_NIPS18_RTNet}           & 75.9                         & -                & -                \\
	NC-Net$_\text{ResNet-101}$~\cite{Rocco_NIPS18_NCNet}         & 78.9                          & 64.7             & 26.4             \\
	DCCNet$_\text{ResNet-101}$~\cite{Huang_ICCV19_DCCNet}       & 82.6                          & 66.1             & 26.7             \\ 
	SFNet$_\text{ResNet-101}$~\cite{Lee2019SFNet}               & 81.9                          & -                & 26.0              \\
	HPF$_\text{ResNet-101}$~\cite{min2019hyperpixel}            & 84.8                         & -                & 28.2             \\
	HPF$_\text{ResNet-101-FCN}$~\cite{min2019hyperpixel}         & \underline{88.3}  & -                & -                \\
	\midrule
	ANC$_\text{ResNet-50}$                                      & 83.7                          & 69.6             & 27.1             \\
	ANC$_\text{ResNet-101}$                                     & 86.1              & \underline{72.4} & \underline{28.7} \\ 
	ANC$_\text{ResNeXt-101}$                                 & \textbf{88.7}    & \textbf{74.1}    & \textbf{30.1}    \\\bottomrule
\end{tabular}
}
\end{table}

\subsection{Benchmark comparisons}
We compare our method with recent state-of-the-art methods, and present our results in Table~\ref{tab:main}.
For results on PF-PASCAL and Spair-71k, all methods are trained on PF-PASCAL.
For results on CUB, the methods are trained and tested on CUB. 
We used three different feature backbones, \ie ResNet-50, ResNet-101, and ResNext-101 for our method.
When using an identical feature backbone (ResNet-101) with other methods, our ANC-Net achieves the best performance on all the datasets.
For example, we achieve $86.1\%$ and $28.7\%$ on PF-PASCAL and Spair-71k respectively.
Note that even with the ResNet-50 feature backbone, our model outperforms NC-Net and DCCNet with the more powerful ResNet-101 feature backbone on all datasets.
Further, when we replace our feature backbone with ResNext-101, the performance of our method can be further boosted on all datasets ($86.1\%$ to $88.7\%$ on PF-PASCAL, $72.4\%$ to $74.1\%$ on CUB, and $28.7\%$ to $30.1\%$ on Spair-71k). Our results are also better than the previous best results achieved HPF with ResNet-101-FCN.
The results clearly demonstrate the effectiveness of our approach. 

\noindent\textbf{Unbiased evaluation on FP-PASCAL.}
As discussed in~\cite{Lee2019SFNet}, there are 302 images in the training split overlapping with either target or source images in the testing split. In terms of images pairs, there are 95 target-to-source pairs in the training split overlapping with the source-to-target pairs in the testing split. Hence, we further conduct an unbiased evaluation by excluding the 302 images and the 95 image pairs respectively. The results are shown in Table~\ref{tab:unbiased}. Our method consistently outperforms NC-Net and DCCNet.

\begin{table}[htb]
\centering
\caption{\textbf{Unbiased evaluation on PF-PASCAL.}}
\label{tab:unbiased}
\resizebox{0.85\linewidth}{!}{
\begin{tabular}{@{}lrrrr@{}}
	\toprule
	Methods                                                 & \multicolumn{1}{c}{Original}        & \multicolumn{1}{c}{w/o 95}  & \multicolumn{1}{c}{w/o 302} \\\midrule\addlinespace[0.5em] 
	NC-Net$_\text{ResNet-101}$~\cite{Rocco_NIPS18_NCNet}                                      & 78.9                          & 78.8             & 80.3             \\
	DCCNet$_\text{ResNet-101}$~\cite{Huang_ICCV19_DCCNet}                                     & 82.6              & 78.7 & 75.7 \\ 
	ANC-Net$_\text{ResNet-101}$                                 & \textbf{86.1}    & \textbf{84.2}    & \textbf{84.5}    \\\bottomrule
\end{tabular}
}
\end{table}

\subsection{Ablation study}
In the ablation experiments, we analyse the effectiveness of all the proposed modules of ANC-Net on PF-PASCAL, with ResNet-101 as the feature backbone.
We experiment on four variants of our ANC-Net, namely, ANC-Net (our model with all components), ANC-Net w/o ANC (our model without ANC, \ie replacing our non-isotropic 4D kernels with the isotropic counterparts), ANC-Net w/o MS (our model with out the multi-scale self-similarity), and ANC-Net w/o Orth (our model trained without orthogonal loss). We also evaluate the three ANC module candidates, denoted as, ANC$_\text{b}$, ANC$_\text{c}$ and ANC$_\text{d}$ in  Figure~\ref{fig:anc}. 
We also compare with NC-Net and DCCNet. 
For a fair comparison with them, we also retrain them with the same sparse annotations.
The retrained NC-Net is the plain baseline of our method, and the retrained DCCNet can be compared with ANC-Net w/o ANC module for evaluating our multi-scale self-similarity module against the self-similarity module of DCCNet. 
The results are reported in Table~\ref{tab:ablation_anc_net}.
As can be seen, when we remove each of our proposed modules, the performance drops, showing that all our proposed modules are effective.
However, ANC-Net and all its variants perform consistently better than the retrained NC-Net and DCCNet as well as the original NC-Net and DCCNet. Among the three ANC architectures in~Figure~\ref{fig:anc},
ANC$_\text{d}$ performs better than the other two by a noticeable margin.
This might be explained by the fact that ANC$_\text{d}$ contains more flexible feature combination paths to deal with objects having more severe scale variations.

\begin{table}[htb]
\centering
\caption{\textbf{Ablation study experimental results.}}
\label{tab:ablation_anc_net}
\resizebox{0.7\linewidth}{!}{
\begin{tabular}{llll}
\toprule
 Method &  PCK@0.1 \\\midrule
NC-Net~\cite{Rocco_NIPS18_NCNet} (original/retrain)       & 78.9/81.9\\
DCCNet~\cite{Huang_ICCV19_DCCNet} (original/retrain) & 82.6/83.7 \\
\midrule

ANC-Net w/o ANC & 84.1 \\
ANC-Net w/o MS   & 84.3\\
ANC-Net w/o Orth & \underline{85.9}\\
ANC-Net w/ ANC$_\text{b}$ &  82.7\\
ANC-Net w/ ANC$_\text{c}$ & 83.8\\
ANC-Net w/ ANC$_\text{d}$  & \textbf{86.1} \\
\bottomrule
\end{tabular}
}
\end{table}


\subsection{Qualitative evaluations}

\begin{figure}[!th]
\centering
  \includegraphics[width=0.85\linewidth]{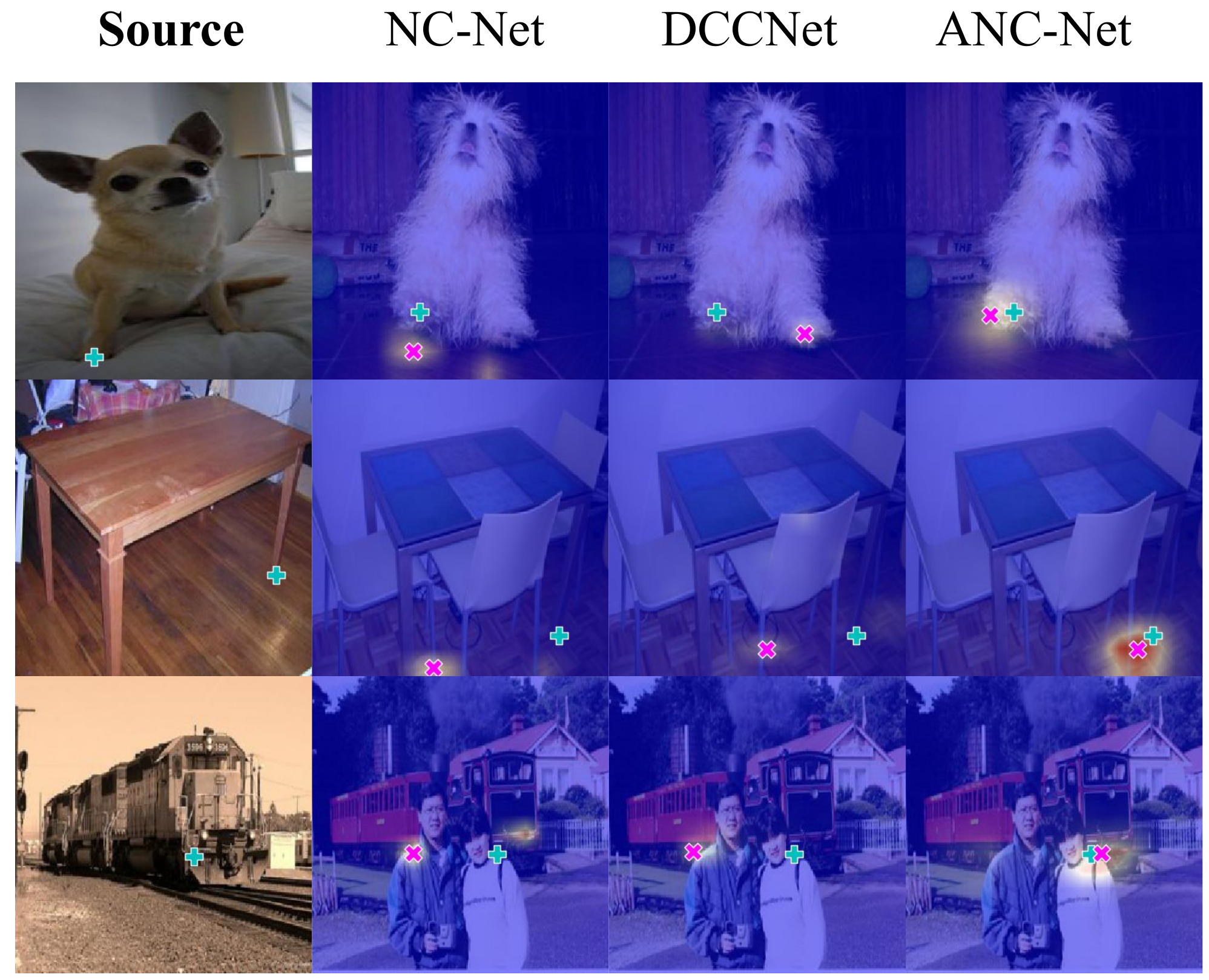}
  \caption{\textbf{Predicted correspondence and correlation map for a query key-point.}
  The first column shows the source images with a query key-point marked with cyan cross. The remaining columns show the correlation maps super-imposed with the target image. 
  The red and cyan crosses represent the prediction and the ground truth respectively.   ANC-Net predicts single-peak correlation maps, avoiding catastrophic failure between distant, but ambiguous key-points, such as the legs of the dog in the first row. 
  Best viewed in electronic form.
  }
\label{fig:orthogonal}
\end{figure}

\begin{figure}[!th]
\centering
  \includegraphics[width=\linewidth]{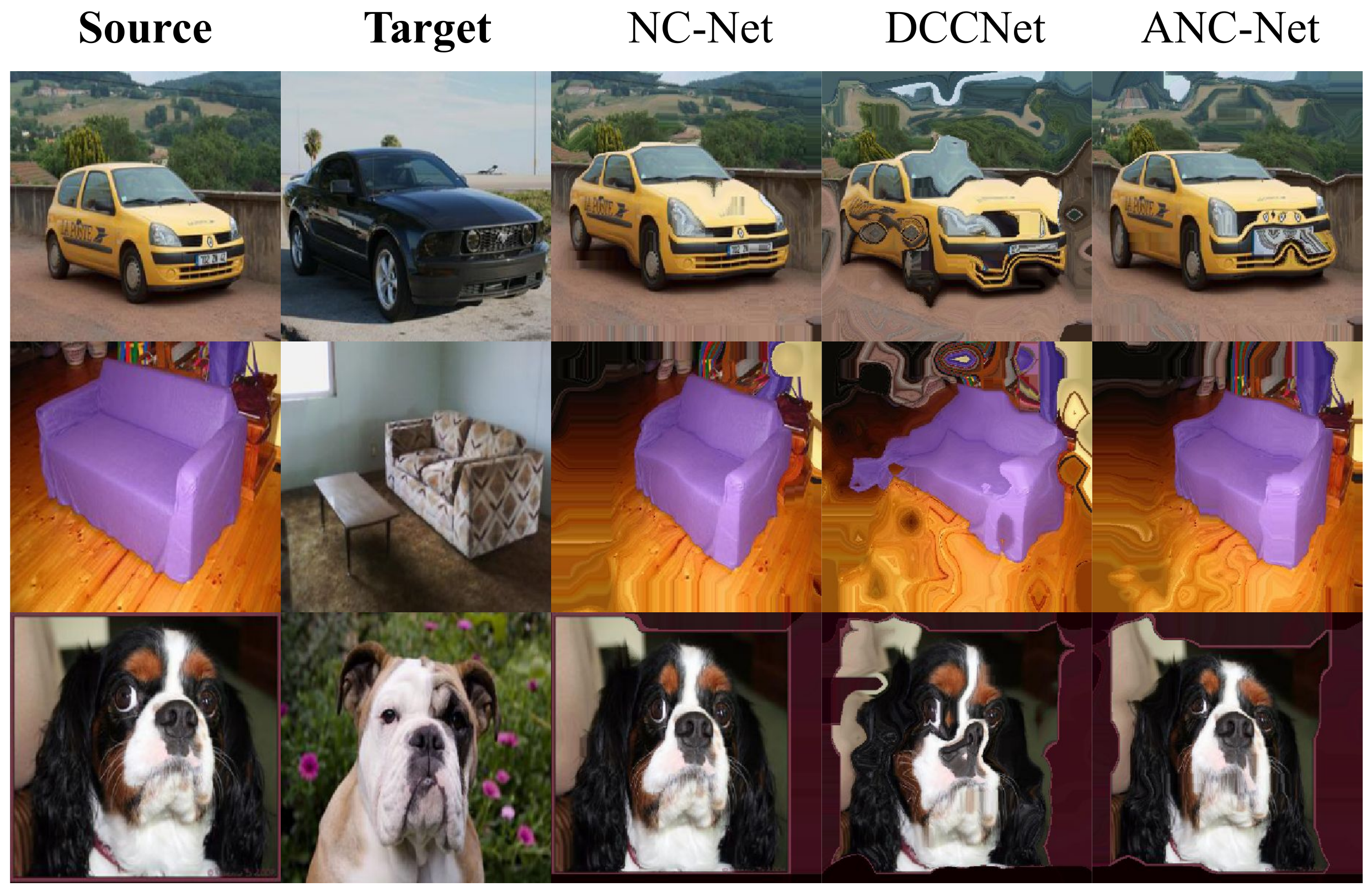}
   \caption{\textbf{Dense correspondence prediction.} Given the correlation map predicted by the model, we compute a dense flow field to warp the source image to the target image. ANC-Net can capture the scale of the objects better than other methods. Best viewed in electronic form.}
\label{fig:warp}
\end{figure}

We show two sets of qualitative experiments. 
The first set of qualitative experiments is shown in Figure~\ref{fig:orthogonal}.
It includes examples of key-points with some degree of ambiguity, such as the limbs of an animal or a table. 
With both NC-Net and DCCNet, it can be seen that there are often multiple peaks in the correlation maps.
In some cases, this can lead to failures where, although the key-points look alike, they are far from the true correspondence.
In contrast, ANC-Net tends to produce correlation maps with a single dominant peak.
This drastically reduces the occurrence of these failures due to the ambiguous nature of a key-point.
We qualitatively evaluate the dense correspondence prediction of ANC-Net in Figure~\ref{fig:warp}.
From a correlation map predicted by the network, we compute a dense flow field, which maps pixel locations from the source to the target image. 
In general, ANC-Net and NC-Net preserve more details in the warping than DCCNet, and ANC-Net is able to capture the scale of the target  more accurately.

%% file: conclusion.tex
\section{Conclusion}

In this paper, we have proposed a convolutional neural network, called ANC-Net, for dense semantic matching.
ANC-Net takes a pair of images depicting different objects from the same category as input, and produces a dense 4D correlation map containing all the pair-wise matches in the feature space.
Pixel-wise semantic correspondences can then be extracted from the 4D correlation map.
ANC-Net can be trained end-to-end with sparse key-point annotations.
At the core of ANC-Net is our proposed 4D non-isotropic convolution kernels, which incorporates an adaptive  neighbourhood consensus constraint for robust matching, and our proposed multi-scale self-similarity module, which aggregates multiple self-similarity features that are insensitive to intra-class appearance variation.
We also proposed a novel loss, called orthogonal loss, that can enforce a one-to-one matching constraint,  encouraging plausible matching results.
We have thoroughly evaluated the effectiveness of our method on various benchmarks, and it substantially outperforms state-of-the-art methods.